\documentclass{article}
\usepackage{graphicx}
\usepackage{tikz}
\usepackage{amsmath}
\usepackage{amsfonts}
\usepackage{xcolor}
\usepackage[capitalize]{cleveref}
\usepackage{subcaption}
\newcommand{\yspace}{0.03\linewidth}

\newcommand{\decompsefigure}[1]{\resizebox{\linewidth}{!}{
\begin{tikzpicture}
  \node(img1) {\includegraphics[width=1\linewidth]{#1}};
  \node(x)[left of=img1, rotate=0, xshift=-0.5\linewidth, yshift=0.27\linewidth] {\large $x_{orig}$};
  \node(psi1)[below of=x, rotate=0, yshift=-\yspace] {\large $\psi_1$};
  \node(psi2)[below of=psi1, rotate=0, yshift=-\yspace] {\large $\psi_2$};
  \node(psi3)[below of=psi2, rotate=0, yshift=-\yspace] {\large $\psi_3$};
  \node(psi4)[below of=psi3, rotate=0, yshift=-\yspace] {\large $\psi_4$};
  \node(psi5)[below of=psi4, rotate=0, yshift=-\yspace] {\large $\psi_5$};
  \draw[red, thick] (-0.5\linewidth,0.11\linewidth) rectangle (0.5\linewidth,0.22\linewidth);
\end{tikzpicture}
}}
\graphicspath{ {./branches_figures/} }

\begin{document}
\title{Additive Class Distinction Maps using Branched-GANs}

\author{Elnatan Kadar \and Jonathan Brokman \and Guy Gilboa}

\maketitle
\begin{abstract}
We present a new model, training procedure and architecture to create precise maps of distinction between two classes of images.
   The objective is to comprehend, in pixel-wise resolution, the unique characteristics of a class.
   These maps can facilitate self-supervised segmentation and object-detection in addition to new capabilities in  explainable AI (XAI).
Our proposed architecture is based on image decomposition, where the output is the sum of multiple generative networks (branched-GANs).  The distinction between classes is isolated in a dedicated branch. This approach allows  clear, precise and interpretable visualization of the unique characteristics of each class. 
We show how our generic method can be used in several modalities for various tasks, such as MRI brain tumor extraction, isolating cars in aerial photography and obtaining feminine and masculine face features. This is a preliminary report of our initial findings and results.

\end{abstract}


\section{Introduction}
In computer vision, identifying the distinction between different classes is a crucial and fundamental task. It is utilized in a wide range of applications, both directly and indirectly. For instance, in an indirect manner, differentiation between classes is essential for classification or detection  {\cite{cai2020review}}. On the other hand, directly identifying the differences between classes is of particular relevance in the field of Explainable AI (XAI)  {\cite{selvaraju2016grad, kohlbrenner2020towards, chatterjee2022weakly}} as it can be used to explain the results of neural network algorithms.
When labeled information that clearly defines the distinction between classes is available, the task becomes relatively straightforward and has been extensively studied   {\cite{zou2023object,gilboa2007nonlocal, wu2022cross, dosovitskiy2020image}}. 
However, a major challenge lies in performing this task in an unsupervised manner, without access to ground truth labels of the differences. In this paper, we focus on an intermediate case in which only general labeling of the classes, such as their identities, is provided, but the unique characteristics or distinctions between the classes are not given. 
This approach is motivated by the fact that for many types of data, it is challenging to obtain detailed pixel-wise accuracy. However, it may be much simpler to achieve general labeling.
A common example is medical imaging, which requires significant knowledge and resources to acquire large amounts of labeled data. Obtaining general labeling through diagnosis, on the other hand, is more manageable. Such labels can be obtained, for instance, by text-processing techniques of the diagnosed scan, without manual intervention. This makes it highly scalable for learning based on large volumes of data.

Having full resolution class distinction maps allows an immediate application of self supervised detection (or even segmentation). In order to train the detection of a certain object, provide a set of images where the object appears and a second set of similar images where the object is absent. The class distinction map would then be the object itself. 
With such a map, relatively simple classical detection algorithms can be used to detect or even to segment the object, without any ground truth.

Our objective is to develop a generic method for creating class distinction maps in a highly reliable manner. 
The maps yield a visual representation of these distinctions, making classification highly interpretable. We propose a general algorithm which does not rely on the characteristics of specific data types  {\cite{wu2021unsupervised, han2021madgan}}. It should perform well on diverse data sets, classes and modalities. Our proposed architecture is based on GANs, but it can be adapted to more advanced generative methods such as diffusion models {\cite{croitoru2022diffusion}} and transformers {\cite{han2022survey}}.

\section{Previous Works}

Numerous studies have been conducted regarding the identification of differences between classes, with a particular emphasis on the subfield of anomaly detection  {\cite{chalapathy2019deep, xia2022gan, salehi2021multiresolution}}. In anomaly detection, one class is deemed anomalous as it contains distinct characteristics that are not present in other classes. The objective of the task is to identify these unique features. While numerous studies have approached this challenge through supervised methods {\cite{salehi2021multiresolution}}, our focus will be on self-supervised techniques, particularly utilizing GANs {\cite{zenati2018adversarially,yang2021lung}}.

The advantage of using GANs in this task lies in its ability to generate realistic images that are similar to those in the dataset. By incorporating an architecture that allows for style transfer  {\cite{choi2020starganv2, deng2022stytr2, lin2020gan}}, we can capture the characteristics of each class in a unique style representation. For example, in medical imaging, we can use style transfer to generate images of tumors from healthy patient images and vice versa, effectively teaching the network what constitutes a tumor. This ability to add or remove features from an image enables the network to obtain precise characteristics of each class.

Many studies rely on cycle loss  {\cite{stepec2021unsupervised,zenati2018adversarially, zia2022vant, you2019ct}} in order to move between classes. Cycle loss assumes  that there is a reversible transformation of transition between the classes.
Other attempts to solve the problem  {\cite{hamghalam2020high, belharbi2021deep}} are by trying to fuse  the anomaly and normal class images in order to generate abnormal images. These works assume that the anomaly is in a specific region of interest in the image, which can be replaced for class transition.
Recent works  {\cite{zia2022vant, nawaz2022mdva}} try to obtain the anomaly by subtraction between the classes .
Here the assumption is that an anomaly image consists of a normal image plus an anomaly. The assumption that the anomaly can be isolated by only two images is simple to implement, although it may be too simplistic in some cases.

Our proposed solution to the problem does not rely on cycle loss and does not assume a reversible transformation between classes, making it computationally more efficient. Unlike other approaches, it does not assume that the differences between classes are localized to specific areas. It adopts a more general perspective of treating an image as a sum of components (simpler images), following \cite{brokman2022analysis}. We show this approach accommodates many scenarios and, although still simple, is sufficiently general.


\section{Method}
\subsection{General branching method}
We propose to utilize a technique known as mixture of experts (MOE)   {\cite{chen2022towards, fedus2022review}} which is a general approach to problem solving. MOE involves breaking down a problem into smaller sub-problems and assigning each sub-problem to an expert. This allows a combination of expert solvers  to provide a general and more robust solution. There are various methods for combining expert solutions within the MOE framework  {\cite{schapire2013explaining}}.
The simplest method is to sum up all proposed solutions. This method is often used for classification or regression problems. We propose to integrate the method also in generation architectures such as GANs {\cite{brokman2022analysis, han2021gan}}. 
That is, to generate several images, each in a separate branch, whose sum is the desired image. In this way we can perform decomposition and analyze each image separately. This allows for a more robust solution, compared to single image generation.

\subsection{Architecture}
\begin{figure}[h!]
    \centering
    \includegraphics[width=1\textwidth]{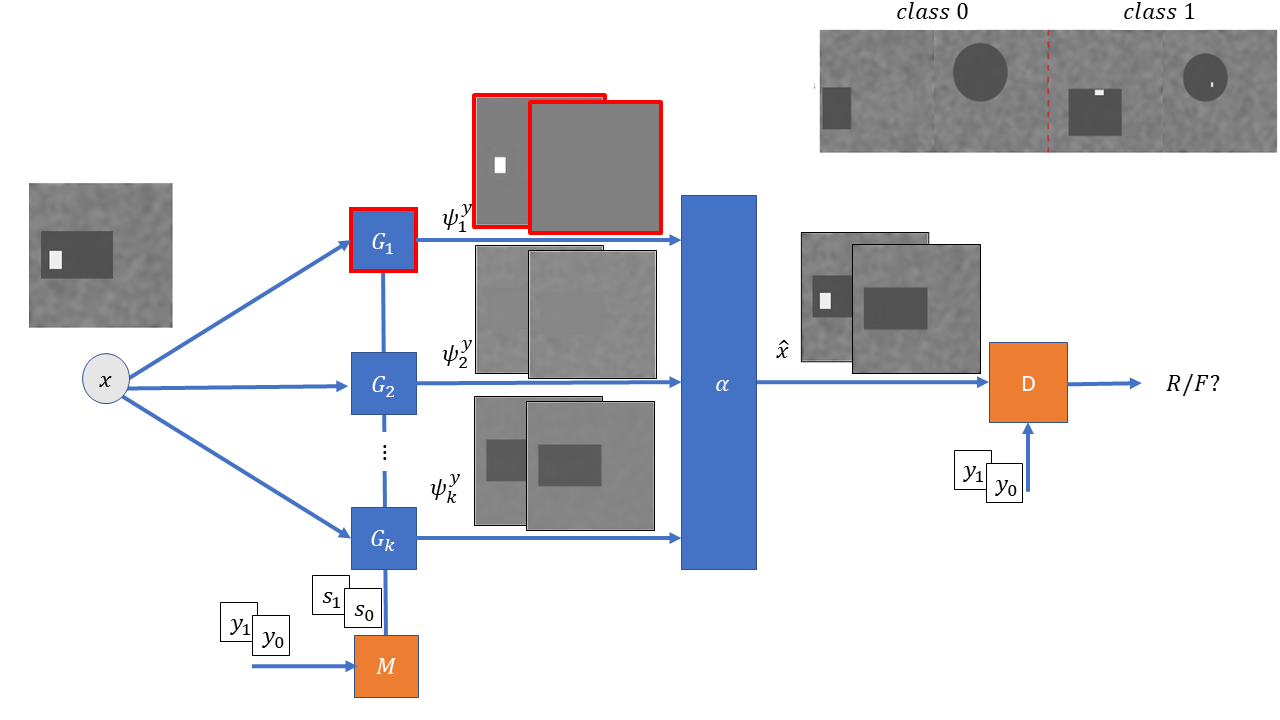}
    \caption{
    During the training phase, each image $x$ is fed into several generators ($G_i$) with the objective of generating two images, one from class 0 and another from class 1. This is achieved by injecting distinct styles into the generators, where the styles are generated through the mapping style ($M$). The goal is to differentiate between the classes and to isolate the differences between them, which is accomplished through the composition process of \cref{eq:alpha_comb} and the loss functions detailed in \cref{Training}. The training process yields two images that are then evaluated by the discriminator, which classifies them based on their class label. Our method produces nearly identical images at the output of all generators, with the exception of $G_1$, which maps the differences between the classes.
    The $\alpha$ block executes a weighted and random superposition process, which effectively separates the information pertaining to the class identity to the first branch, while excluding the information irrelevant to the class identity in the other branches (Details in \cref{Training}). 
    }
    \label{fig:training_model}
\end{figure}

\begin{figure}[h]
    \centering
    \includegraphics[width=1\textwidth]{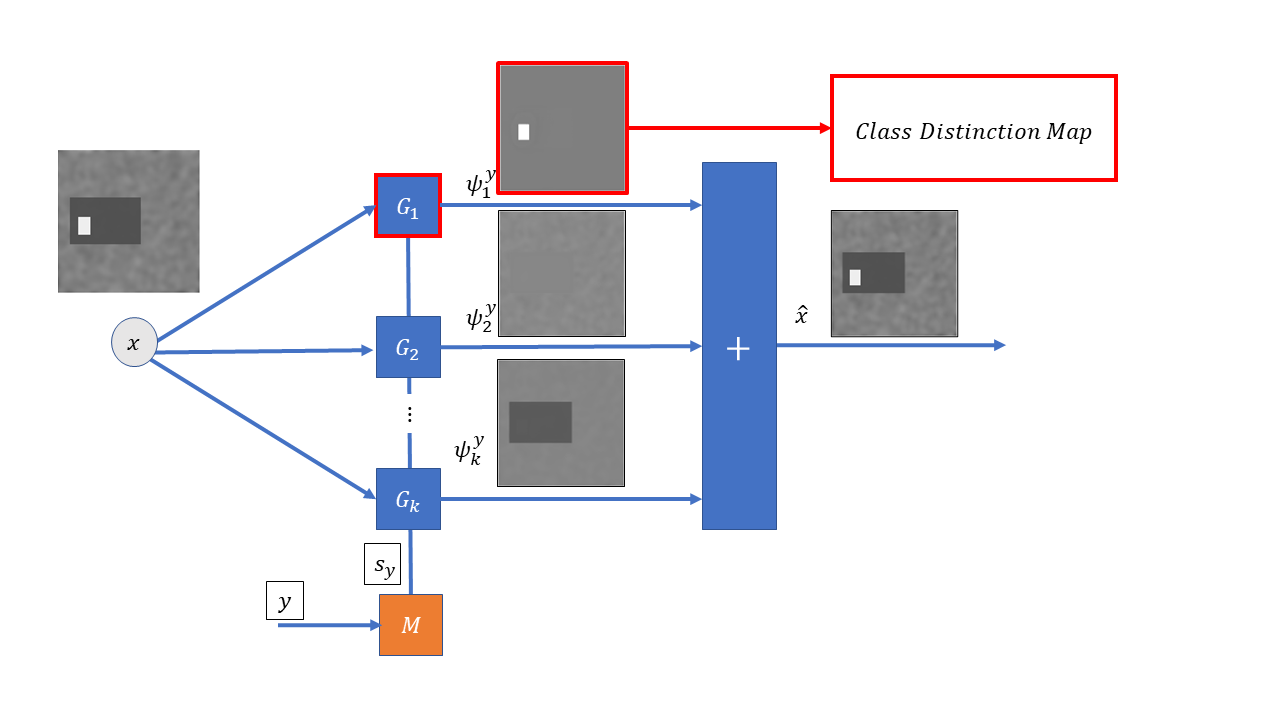}
    \caption{
    Inference stage: We assume to have an image and a class as inputs.
    The image is fed into the generators with a style appropriate for its class. The generators perform decomposition, resulting in the creation of several images with varying characteristics, which are then combined to form the generated image, which should be as close as possible to the original image. The class distinction map is produced in the first branch.}
    \label{fig:detecting_model}
\end{figure}

We adopt the StarGAN-v2 architecture \cite{choi2020starganv2} as our base model for image-to-image translation. Several modifications have been made to the original architecture in order to adapt it to our task. The three most significant changes are as follows:
\begin{itemize}
\item {\bf Removal of the style-encoder component.} Instead of using the style encoder to extract style information from a specific image, we propose to use only mapping-style  to create style vectors that are based solely on the identity of the class.
\item {\bf Splitting the generator into $N$ generators.}  Decomposing the generator into multiple sub-generators, each responsible for generating a specific aspect of the image. The outputs of these sub-generators are then summed together to obtain the overall image, which is then fed into the discriminator during training \cite{brokman2022analysis}. See
\cref{fig:training_model} and \cref{fig:detecting_model}.
\item {\bf Addition of skip connections.} Similar to the U-net architecture \cite{ronneberger2015u}, we propose to add skip connections between the encoder and decoder layers. This is done to preserve delicate details of the image in high resolution.
\end{itemize}
\subsection{Assumptions and Notations}

Our approach assumes that each image can be expressed as a composite of various sub-images. Some of these sub-images contain class-identifying information, while others do not. The objective is to aggregate all sub-images that are instrumental in determining the class identity into a single image. 

In the context of this task, labeled information is only given at an image level (and not at a pixel resolution). Specifically, the distinctions or variations between the images are not annotated. The data provided consists of a set of images belonging to two distinct classes, with general class labels.
We annotate the training set as X, which consists of two sets of images (with a similar size): $X=\{X^0,X^1\}$, where $X^y$ belong to class $y=\{0,1\}$.
We denote the output of each generator as
\begin{equation}
    \psi_i = G_i(x,s).
\end{equation}
The generated image is the following sum,
\begin{equation}
\hat{x} = \sum\limits_{i=1}^N \psi_i .
\end{equation}
In particular, we assume that \(\psi_1\) contains the class-identifying information, while others do not.
Therefore, we can split the sum into two parts:
\begin{equation}
\hat{x} =\psi_1+\sum\limits_{i=2}^N \psi_i .
\end{equation}
 We decompose the output image into two components: one is unique to a specific class and the other comprises of all common information relevant to both classes. By utilizing a mixture of specialized branches and allowing each branch to distinguish different characteristics, we are able to achieve a more robust reconstruction. As demonstrated in \cref{fig:toy_PSNR} and \cref{fig:psnr_num_branches}, a branched approach yields a higher quality reconstruction, compared to a single generator methods.
 
{\bf Additional notations.} An image from the data belonging to class \(y\) is denoted by \(x^y\). An image that is generated by the style of class \(y\) is
 \(\hat{x}^y=\sum\limits_{i=1}^N \psi_i^y\), where
\(\psi_i^y=G_i(x,s_y)\).
An image undergone a style transfer process is denoted by $x^{st}$. This image is generated based on the complementary class of $x$. Conversely, an image that has not undergone style transfer and is generated based on its actual class is denoted by $x^{\Tilde{st}}$. It is expected that $x^{\Tilde{st}} \approx x$ with minimal levels of distortions.

\subsection{Training} \label{Training}
During training, given an image $x^y$ from the training set, two images are generated (by the group of generators) of the same class as $x^y$ and of the complementary class, $\hat{x}^0$ and $\hat{x}^1$.
The objective is to generate an image similar to the input image while incorporating the attributes of the target class. In case the target class is identical to the original class, the generated image is expected to remain unchanged, i.e., $\hat{x}^y \approx x^y$.

\subsubsection{New $\alpha$-blended generation}
We would like to have the first channel contain the class distinction information and the rest of the channels to contain information common to both classes.  We thus propose a novel $\alpha$-blending mechanism. 
 For each batch, we generate a random vector $\alpha$ of length \(N-1\) uniformly distributed between $0$ and $1$. The images are then generated during the training as follows:
\begin{equation}
\begin{aligned}
    \hat{x}^0=\psi_1^0+\sum\limits_{i=2}^N \alpha_{i-1}\psi_i^0+\sum\limits_{i=2}^N (1-\alpha_{i-1})\psi_i^1, \\
    \hat{x}^1=\psi_1^1+\sum\limits_{i=2}^N \alpha_{i-1}\psi_i^1+\sum\limits_{i=2}^N (1-\alpha_{i-1})\psi_i^0.
\end{aligned}
\label{eq:alpha_comb}
\end{equation}
The proposed method encourages the generators to generate identical images for both branches in the sum and thus isolate the distinction between the classes to $\psi_1$. In the ideal case, where the branches in the sum are identical and the distinction is only in $\psi_1$ the expression of each branch converges to $\hat{x}^0=\psi_1^0+\sum\limits_{i=2}^N\psi_i^0$.
This is the way we will generate the image in the inference stage.
To obtain the desired results, a set of requirements must be imposed, which are defined through the use of the several loss functions.

\subsubsection{Reconstruction losses}
When we generate an image from the same class as the input we would like both images to be very similar. 
We use both $L^1$-norm and $L^2$-norm fidelity terms, \cref{eq:L_rec}.
In addition, to be highly faithful near edges, we also require a good reconstruction of the gradients, \cref{eq:grad-rec}. This is inspired by \cite{sitzmann2020implicit}.
Moreover, our observations indicate that regions in which the dissimilarity between classes is pronounced pose greater challenges for our model in reconstruction results. As such, we placed more emphasis on ensuring similarity in regions where significant differences were detected (\cref{eq:dis-rec}), i.e., those characterized by a large amplitude in the first branch. We define large amplitude as an absolute value exceeding the average of the absolute values observed throughout the entire branch:
\begin{equation}
L_{rec}(x,\hat{x}) = \sum_{over-pixels} \left[|x-\hat{x}|+(x-\hat{x})^2\right]
\label{eq:L_rec}
\end{equation}

\begin{equation}
L_{grad-rec}(x,\hat{x}) = L_{rec}(\nabla x,\nabla\hat{x})
\label{eq:grad-rec}
\end{equation}

\begin{equation}
    \begin{aligned}
        L_{dis-rec}(x,\hat{x})&=\frac{L_{rec}(x \odot \mathbb{I}, \hat{x}^{\tilde{st}} \odot \mathbb{I})}{\sum \mathbb{I}} , 
        \\ 
        \mathbb{I}&= 
        \begin{cases} 
            1 & \text{if $|\psi_1^{\Tilde{st}}|>mean(|\psi_1^{\Tilde{st}}|)$} \\
            0  & \text{else},
        \end{cases}
    \end{aligned}
    \label{eq:dis-rec}
\end{equation}
$\odot$ is a notation for element-wise product.
\subsubsection{Specialization losses}
We use multiple branches, where each branch specializes in distinct characteristics, this improves training speed and reduces reconstruction error. We would like to ensure that information in one branch has little overlap with information in other branches. To accomplish this, we propose the following losses,
\begin{equation}
    L_{sqr}(\psi_1,...,\psi_N) =\left\Vert\sum\limits_{i=1}^N \psi_i^2-\left(\sum\limits_{i=1}^N \psi_i\right)^2\right\Vert^2,
\label{eq:sqr_loss}
\end{equation}

\cref{eq:sqr_loss} promotes that for each pixel, there is a single dominant branch contributing its value to the output image while the rest of the branches contribute approximately zero. To prevent the occurrence of the degenerate scenario where all the information is produced in a single branch 
we use a regularization objective which promotes equal energies in the non-class  branches.

\subsubsection{Sparsity losses}
Due to the summation of multiple branches in generating the images (where the values range from -1 to 1), redundant information results in undesired amplitude in certain areas. To address this issue, we aim to promote sparsity in the first branch, which carries the class distinction information, so that it only contains relevant information with minimal noise.
We use the following loss:

\begin{equation}
    L_{sparse}(\psi_1) = \left\Vert\psi_1\right\Vert_1.
    \label{eq:L_sparse}
\end{equation}
which is a standard $L_1$ sparsity loss.

All the loss functions mentioned above come in addition to the adversarial loss $L_{adv}$ listed in \cite{choi2020starganv2}. To train the generators, we need to minimize the overall objective, which is:
\begin{equation}
    \begin{split}
        L(G_1, ..., G_N) = \lambda_{rec}L_{rec} +
        \lambda_{grad-rec}L_{grad-rec} +\lambda_{dis-rec}L_{dis-rec}
        \\+\lambda_{sqr}L_{sqr}
        +\lambda_{sparse}L_{sparse}
        + \lambda_{adv}L_{adv}.
        \label{eq:L_total}
    \end{split}
\end{equation}
The discriminator is trained identically to \cite{choi2020starganv2}.

\section{Experimental Results}

\subsection{Toy Examples}
In order to effectively present the concept of our proposed method, we first demonstrate it using synthetic data.
Images in this dataset possess three distinct features: a background composed of uniformly distributed Gaussian-smoothed noise;
a dark shape in the form of either a rectangle or a circle, with random size and at random position;
a white rectangle of varying size and position that may or may not be present within the dark shape.

\begin{figure}[h]
    \centering
    \includegraphics[width=\textwidth]{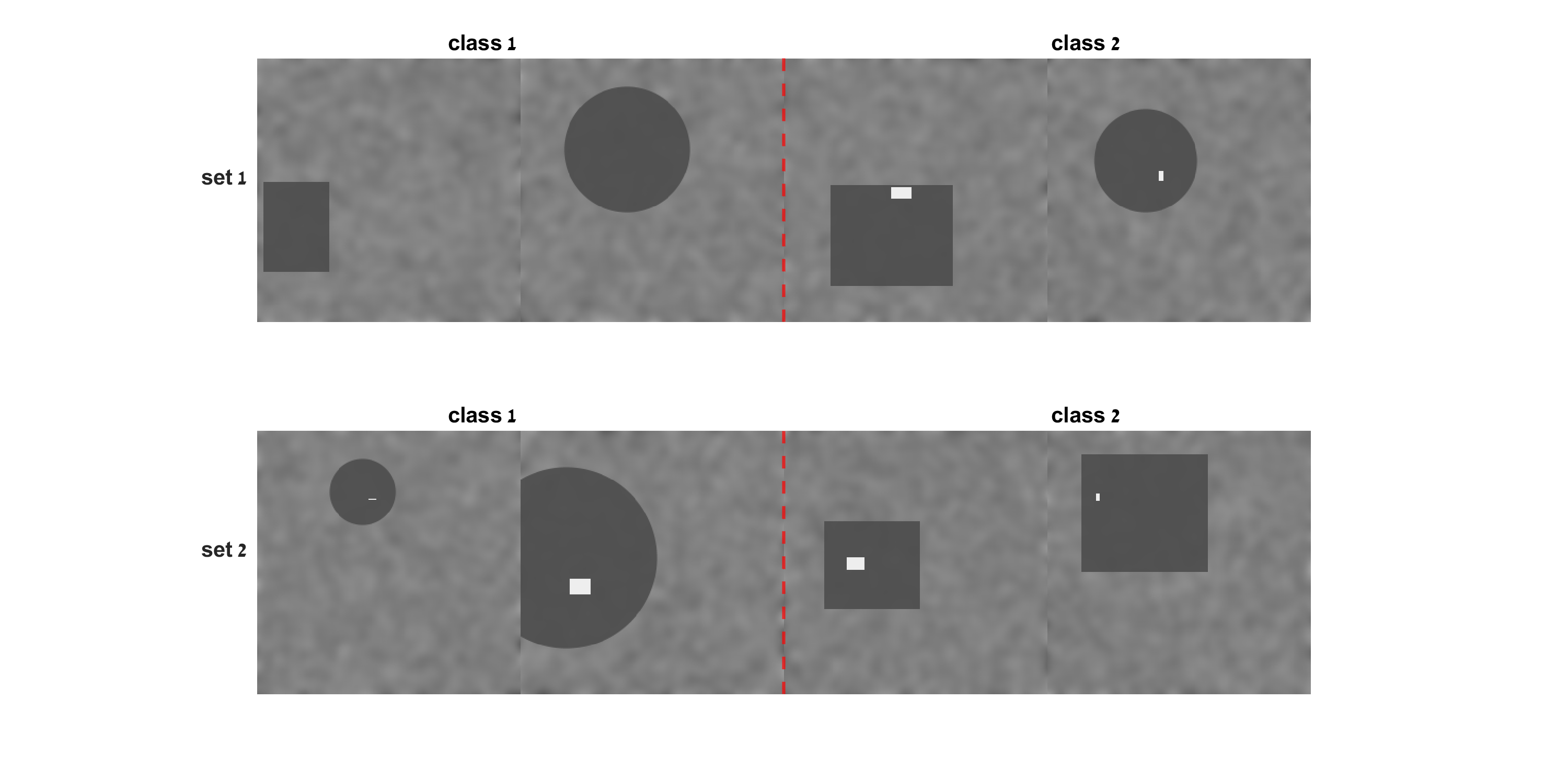}
    \caption{An illustration of the division of synthetic data into two classes using two different methods. The first set demonstrates divisions based on the presence or absence of a white rectangle, while the second set illustrates divisions based on the shape of the dark feature (circle or rectangle).}
    \label{fig:display_classes_toy}
\end{figure}

We have implemented two distinct forms of division. In the first, images are separated into two groups based on the presence or absence of a white rectangle within the dark shape. In the second form, images are divided into two groups based on the shape of the dark feature (circle or rectangle). 
An example of these divisions is shown in \cref{fig:display_classes_toy}.

\begin{figure}[h!]
    \begin{subfigure}{0.5\textwidth}
    \includegraphics[width=1\textwidth]{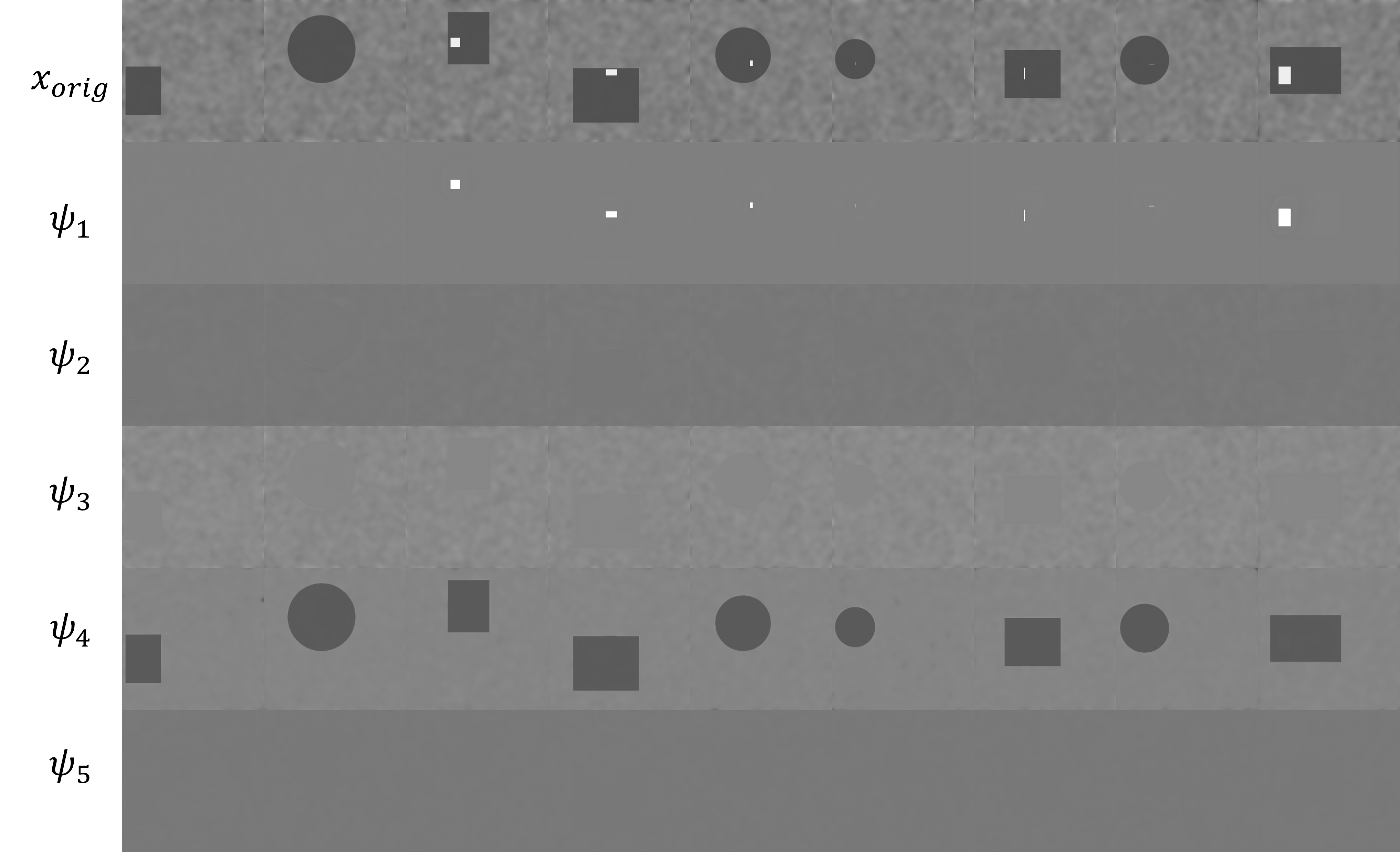}
        \subcaption{set 1}
    \end{subfigure}
    \begin{subfigure}{0.5\textwidth}
    \includegraphics[width=1\textwidth]{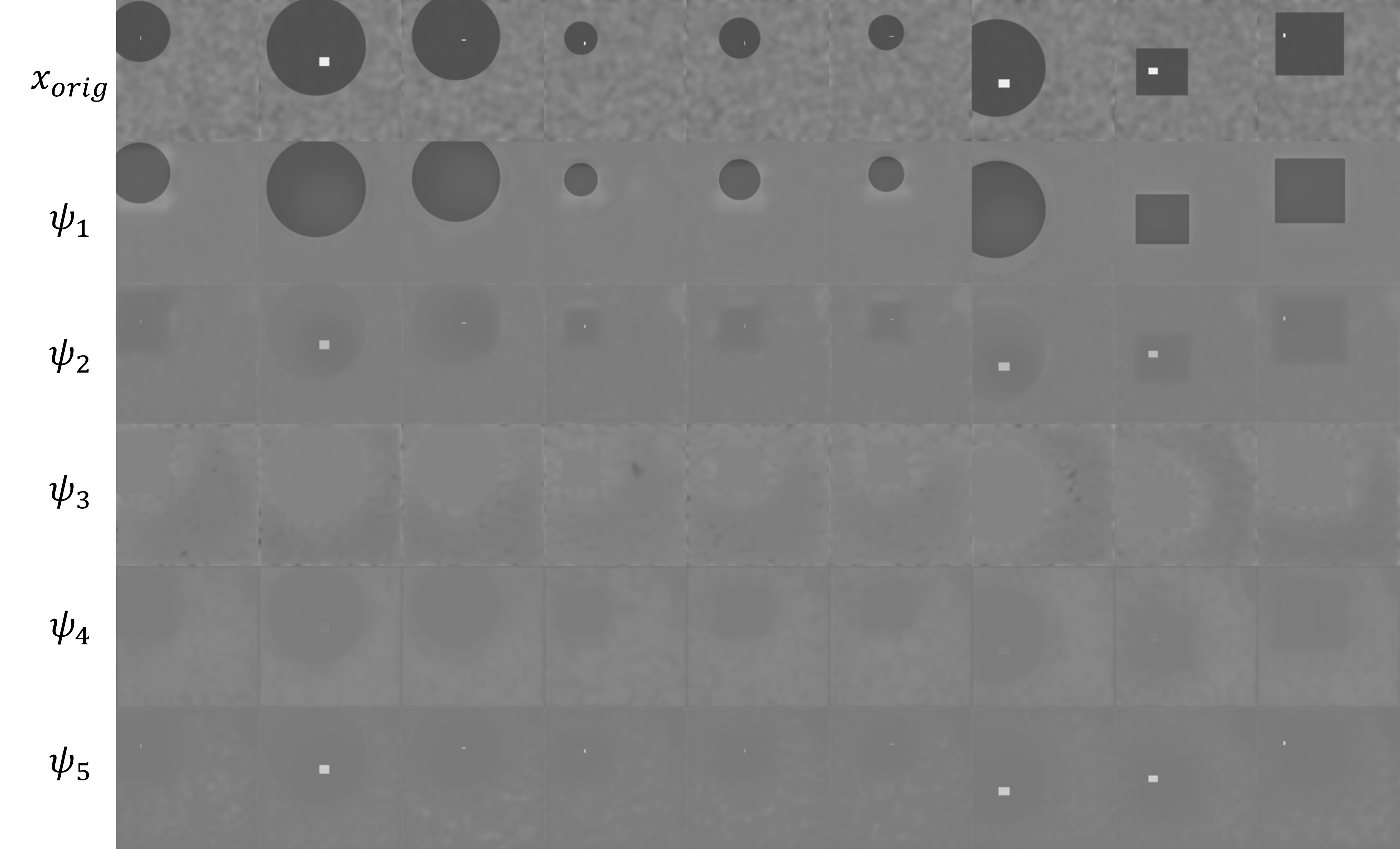}
        \subcaption{set 2}
    \end{subfigure}
     \caption{{decomposition examples:} The first line in each image shows the original input. Each row has the output of a different branch. The first branch contains the difference between the classes and is marked with the red rectangle. (a) set 1 (b) set 2, which appear in ,\cref{fig:display_classes_toy}}
    \label{fig:toy_anomaly}
\end{figure}

We expect to obtain in the first branch the distinction between the two groups, whereas in the remaining branches we get decompositions to components that are less relevant to class identity. The results of this experiment can be seen in the accompanying \cref{fig:toy_anomaly}
. We clearly observe that the first branch isolates well the features which distinct the two classes for both very different forms of division. Specifically, for set 1, the presence of a white rectangle is highlighted, whereas for set 2, the dark shape is emphasized. This demonstrates that the algorithm was able to successfully distinguish between the two classes, and the results can be easily interpreted through visual inspection.

\begin{figure}[h!]
    \begin{subfigure}{0.5\textwidth}
    \centering
    \includegraphics[width=1\textwidth]{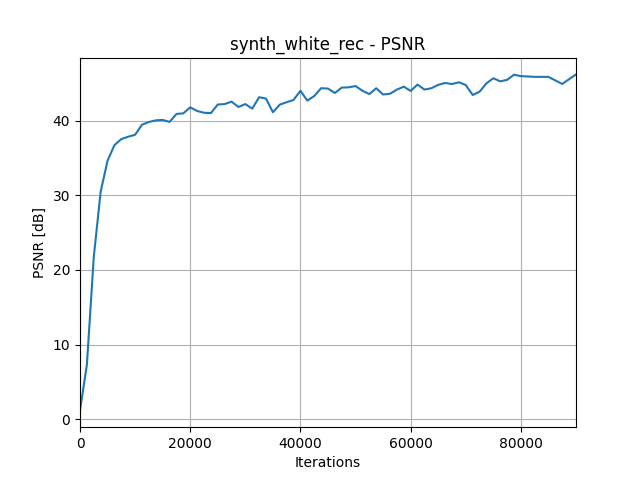}
    \caption{}
    \end{subfigure}
    \hfill
    \begin{subfigure}{0.5\textwidth}
    \centering
    \includegraphics[width=1\textwidth]{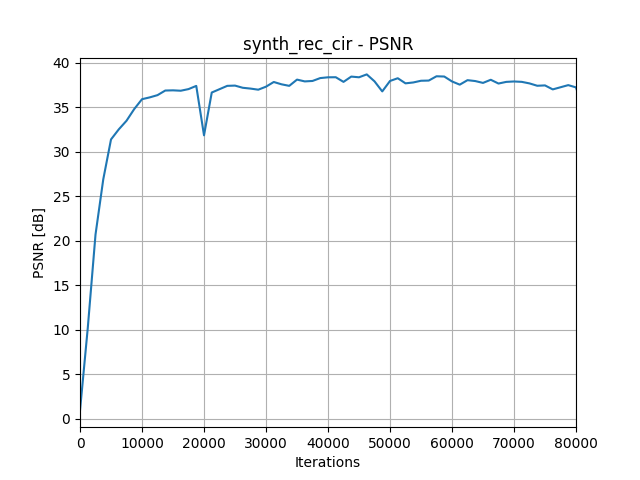}
    \caption{}
    \end{subfigure}
    \caption{(a) set 1 PSNR. (b) set 2 PSNR.}
    \label{fig:toy_PSNR}
\end{figure}

Another important issue, as discussed earlier, is the aim to preserve the integrity of the input image during  decomposition, avoiding distortions. At inference, using the correct input class, we would like to receive a reconstruction very similar to the input image. 
To quantify the reconstruction quality, we compute the average Peak Signal-to-Noise Ratio (PSNR) of the reconstructed images for the test set. As shown in \cref{fig:toy_PSNR}, the PSNR values indicate a reconstruction of very high quality.

\subsection{Anomaly Detection}
\begin{figure}[h!]
\decompsefigure{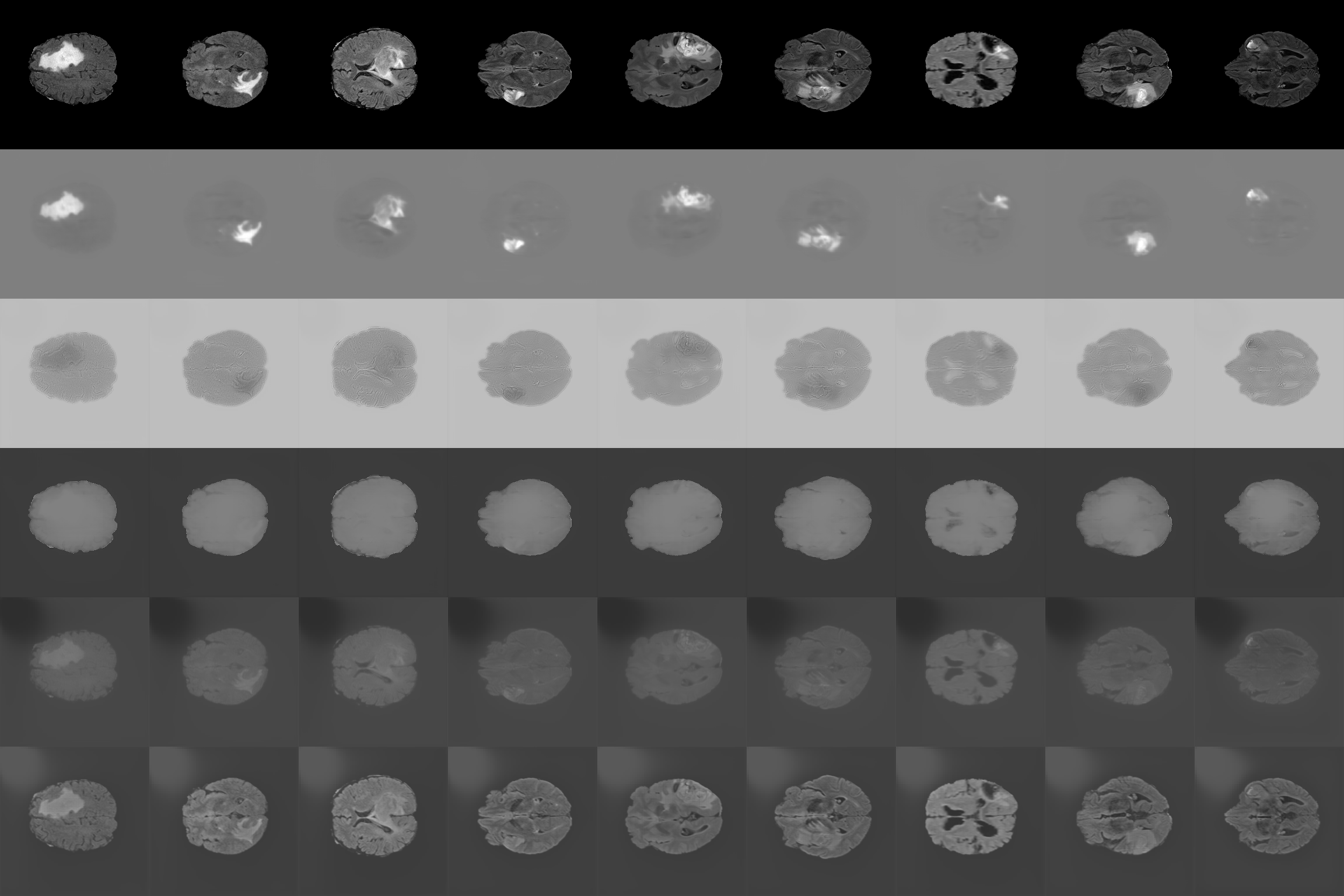}
\caption{Finding tumors results}
\label{fig:BraTS_decomposition}
\end{figure}

We now evaluate the performance of our proposed algorithm for the task of anomaly detection. This task is a special case of identifying differences between classes, with one class being defined as normal and the other as abnormal. The objective is to identify the factor that causes the image to be classified as abnormal. This task is commonly utilized for detecting abnormalities in medical imaging, such as tumors or fractures in MRI, CT or X-ray scans. In these scenarios, an image containing a pathology would be considered abnormal, in comparison to a healthy image (normal).
To evaluate the algorithm's performance in this task, we utilized the BraTS dataset \cite{menze2014multimodal}. It contains magnetic resonance imaging (MRI) scans of the brain with and without tumors. This dataset is widely used in the literature to demonstrate the performance of various methods in detecting tumors. However, it is important to note that we did not utilize the labeled information of the tumors themselves, but rather identified them based on the general label of the presence or absence of a tumor in the image. Therefore, we expect to observe tumors in the first branch \ref{fig:BraTS_decomposition}, which contains the information on the identity of the class, when comparing images with and without tumors.

\subsection{Identifying cars in aerial photographs}
\begin{figure}[h!]
\decompsefigure{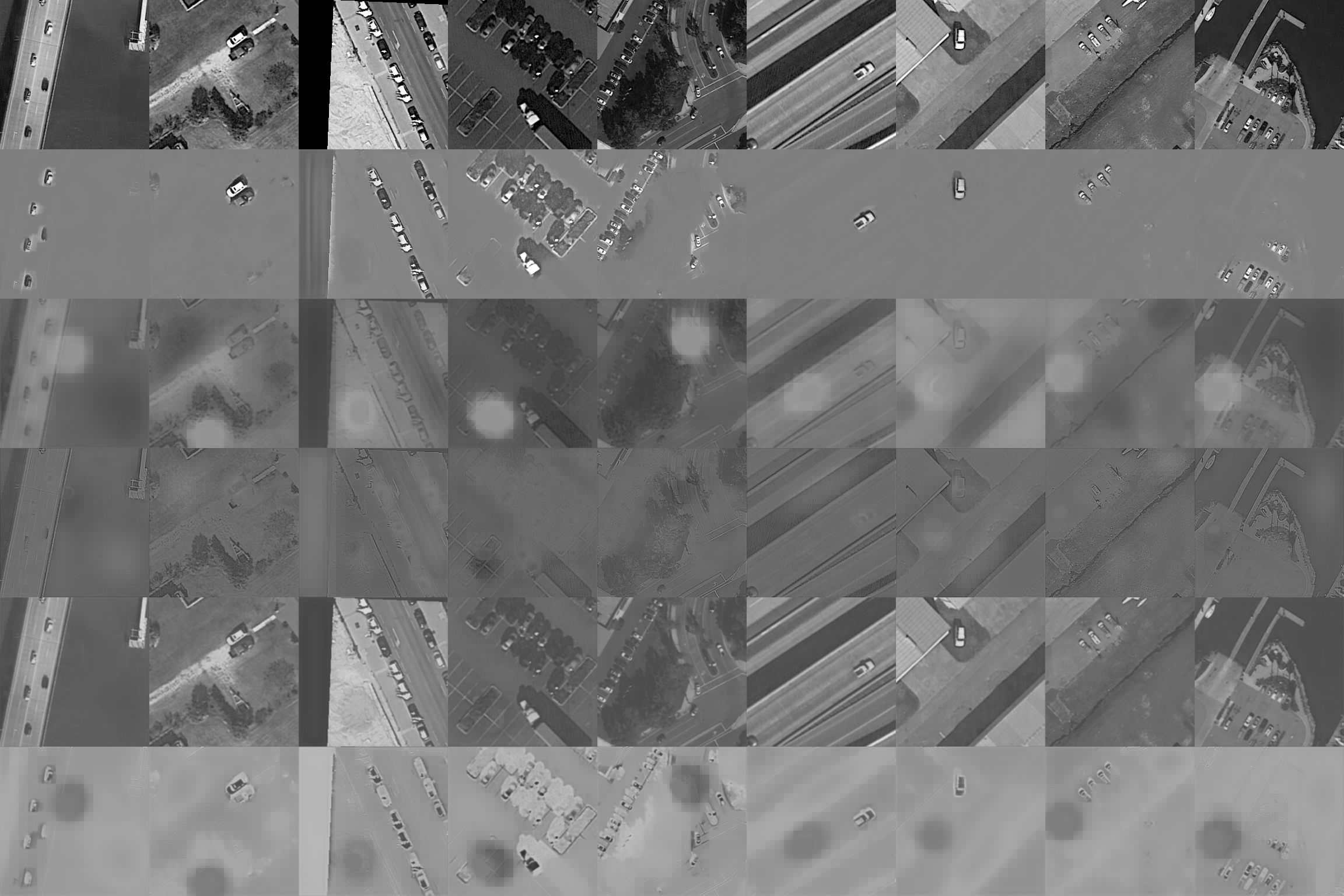}
\caption{Finding cars results}
\label{fig:DOTA_cars_decomposition}
\end{figure}

\begin{figure}[h!]
\decompsefigure{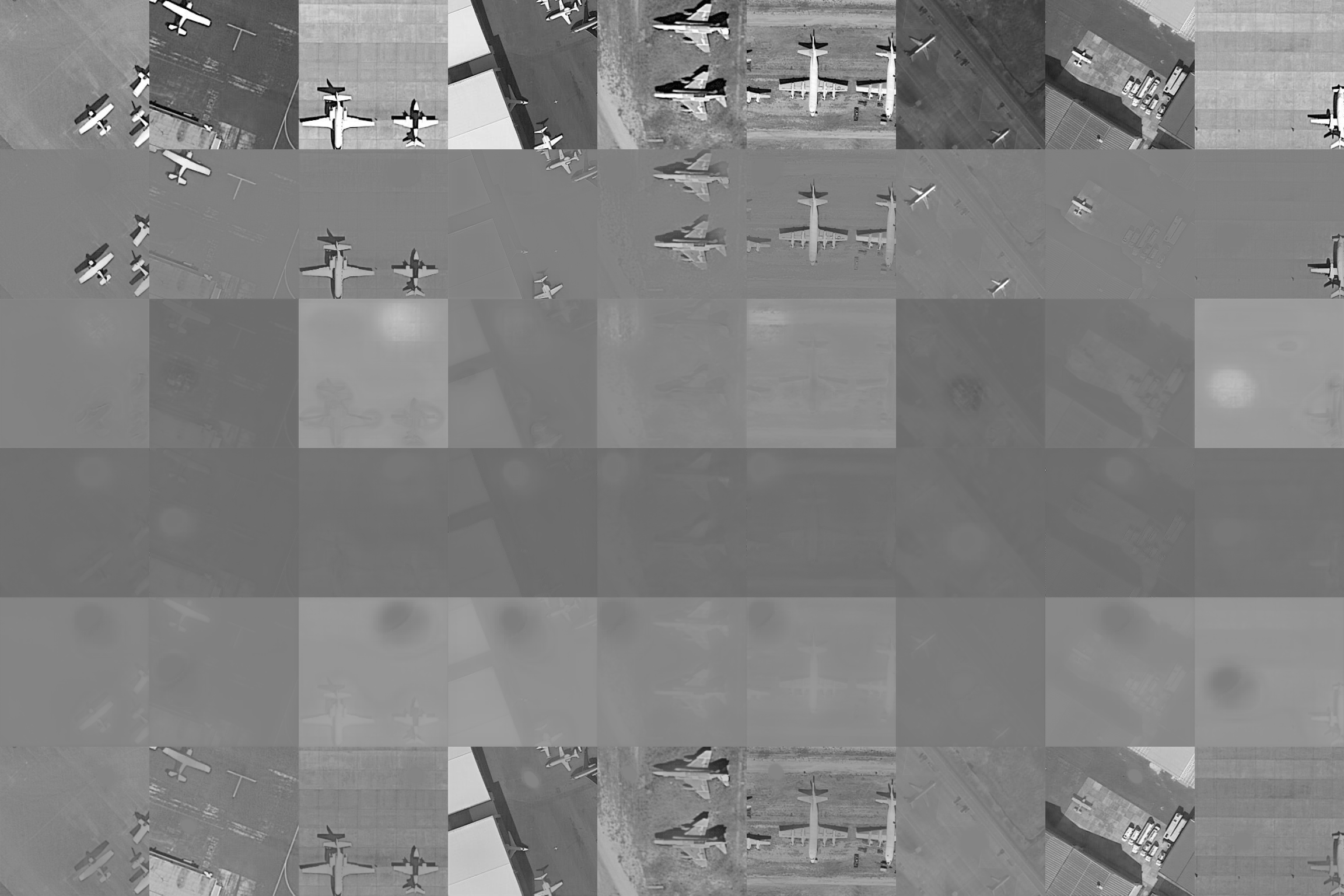}
\caption{Finding planes results}
\label{fig:DOTA_planes_decomposition}
\end{figure}

In another experiment, we applied the algorithm to address a problem in aerial photography. Specifically, our aim was to identify cars in high-resolution aerial photographs from the DOTA dataset \cite{xia2018dota} without prior annotation of car locations. To accomplish this, we extracted image patches of size 256x256 from the dataset and divided them into two groups: one containing images with cars and one without. Given the distinct characteristics of the two classes, the class distinction map should contain information that could be used to extract the location of the cars. Our experimental results,  \cref{fig:DOTA_cars_decomposition}, show a successful isolation of mostly the car areas in the image. This seems as a promising approach to be used for self-supervised detection and segmentation methods in aerial photography. It can be done without the need for manual annotation, which could have significant implications for tagging aerial images that are traditionally difficult to annotate due to their high level of detail. Likewise, our algorithm has also successfully distinguished planes, \cref{fig:DOTA_planes_decomposition}, through a different categorization of images into two classes, one containing planes and the other not. In this case as well, the location of the planes was not explicitly used.

\subsection{Highlighting gender differences}
\begin{figure}[h!]
\decompsefigure{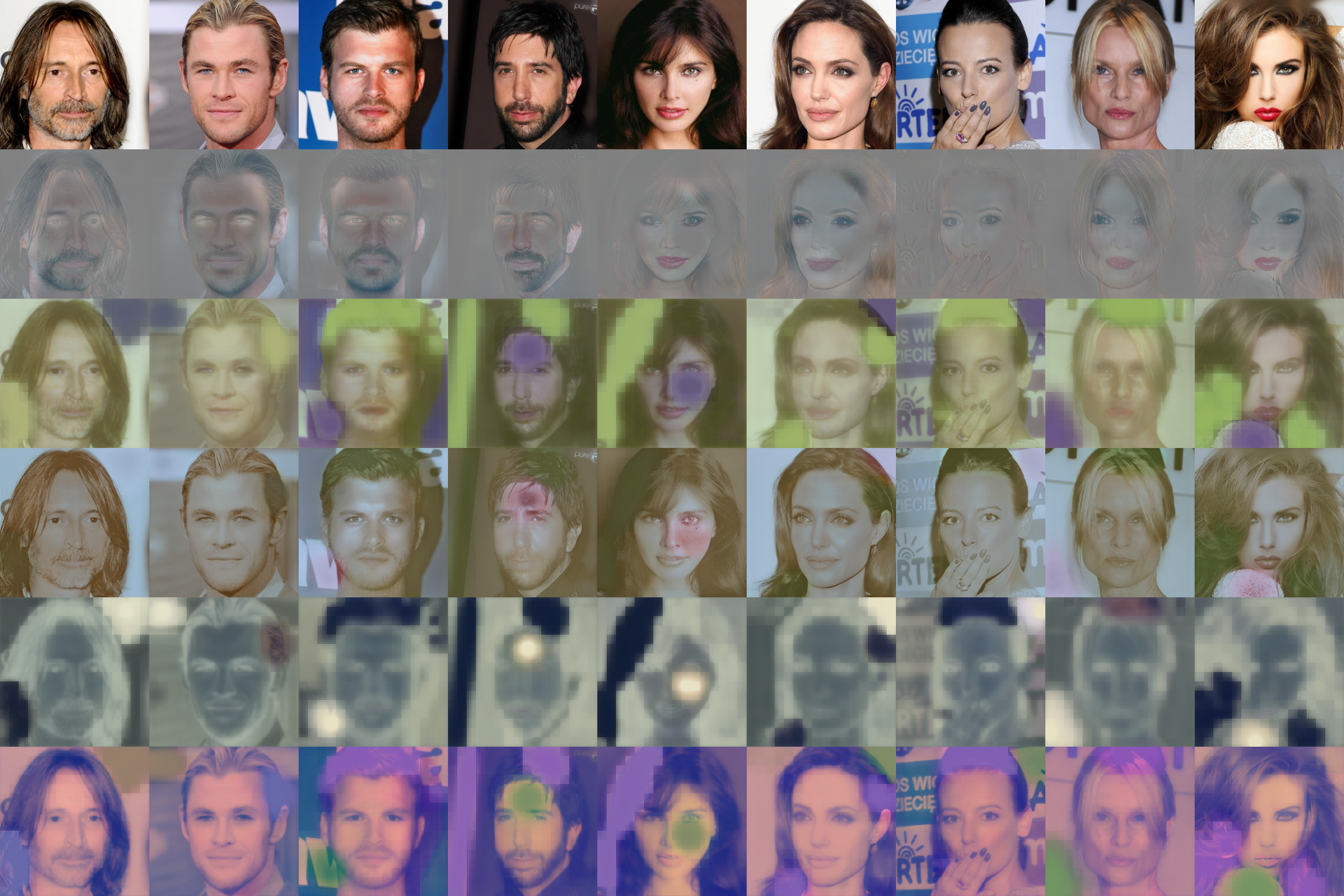}
\caption{Decomposition of CelebA. The first branch of the decomposition carries information about the class identity. Specifically, for men's photos, the branch emphasizes the presence of facial hair, eyebrows, and related features, while for women's photos, make-up on the lips and eyes is highlighted.}
\label{fig:celeba_decomposition}
\end{figure}

We also examined our algorithm on the CelebA face image dataset\cite{liu2015faceattributes}. Specifically, we divided the dataset into two classes: men and women, and aimed to determine if the algorithm was able to isolate and highlight characteristics that differentiate the two classes. Our results (\cref{fig:celeba_decomposition}) show clear distinctions of the genders. For example, in the female class, the algorithm emphasized makeup, hair, and jewelry, while in the male class, the algorithm emphasized beards, eyebrows, and hair. These results suggest that our general method is able to effectively explain differences between classes in a visually interpretable manner.

\subsection{Ablation: influence of branches number}

\begin{figure}[h!]
    \begin{subfigure}[b]{0.5\textwidth}
    \centering
    \includegraphics[width=\textwidth]{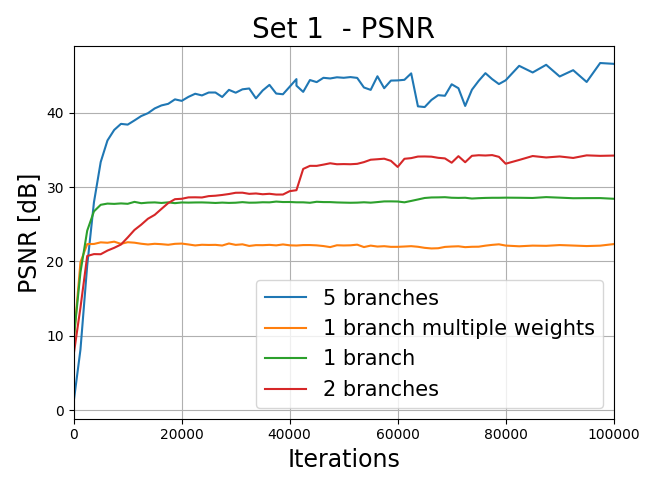}
    \caption{}
    \end{subfigure}
    \hfill
    \begin{subfigure}[b]{0.5\textwidth}
    \centering
    \includegraphics[width=\textwidth]{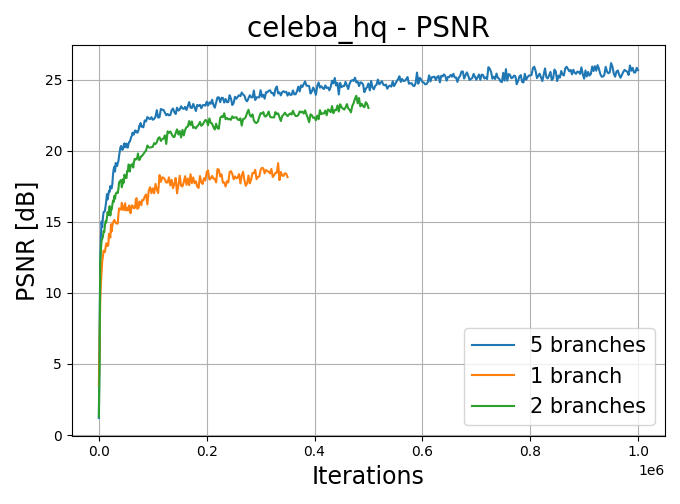}
    \caption{}
    \end{subfigure}
    \caption{(a) set 1 PSNR. (b) Celeb A PSNR.}
    \label{fig:psnr_num_branches}
\end{figure}

In order to justify the use of multiple branches, we conducted experiments to assess the performance for various branch configurations. Specifically, we tested the use of one branch, two branches, and five branches. A single branch represents a degenerate case of style transfer, while the two-branch configuration simulates methods such as VANT-GAN \cite{zia2022vant}, and the five-branch configuration incorporates the advantages of our proposed method. As can be seen in \cref{fig:psnr_num_branches}, more branches yield better performance, in terms of PSNR.

In order to rule out the possibility that the improved PSNR was solely due to the increased number of network parameters, we also conducted an experiment with a single large branch (number of parameters similar to the five-branch configuration). The results show that the performance was even worse, suggesting that a larger number of branches does indeed contribute to the reconstruction quality. We believe this improvement is likely due to the MOE phenomenon, as each branch specializes in different characteristics, thereby complementing one another and resulting in a better overall reconstruction.

\section{conclusion}
This work presents a novel method for detecting differences between classes using a GAN architecture. The results demonstrate the generality and robustness of the method. It is applicable to a wide range of data types and produces visually interpretable outputs for various tasks. We show here preliminary results of this promising direction and intend to further explore the method with additional generalizations and uses. 

\begin{figure}[h!]
    \resizebox{\textwidth}{!}{
    \begin{tikzpicture}
        \node(img1) {\includegraphics[width=1\textwidth]{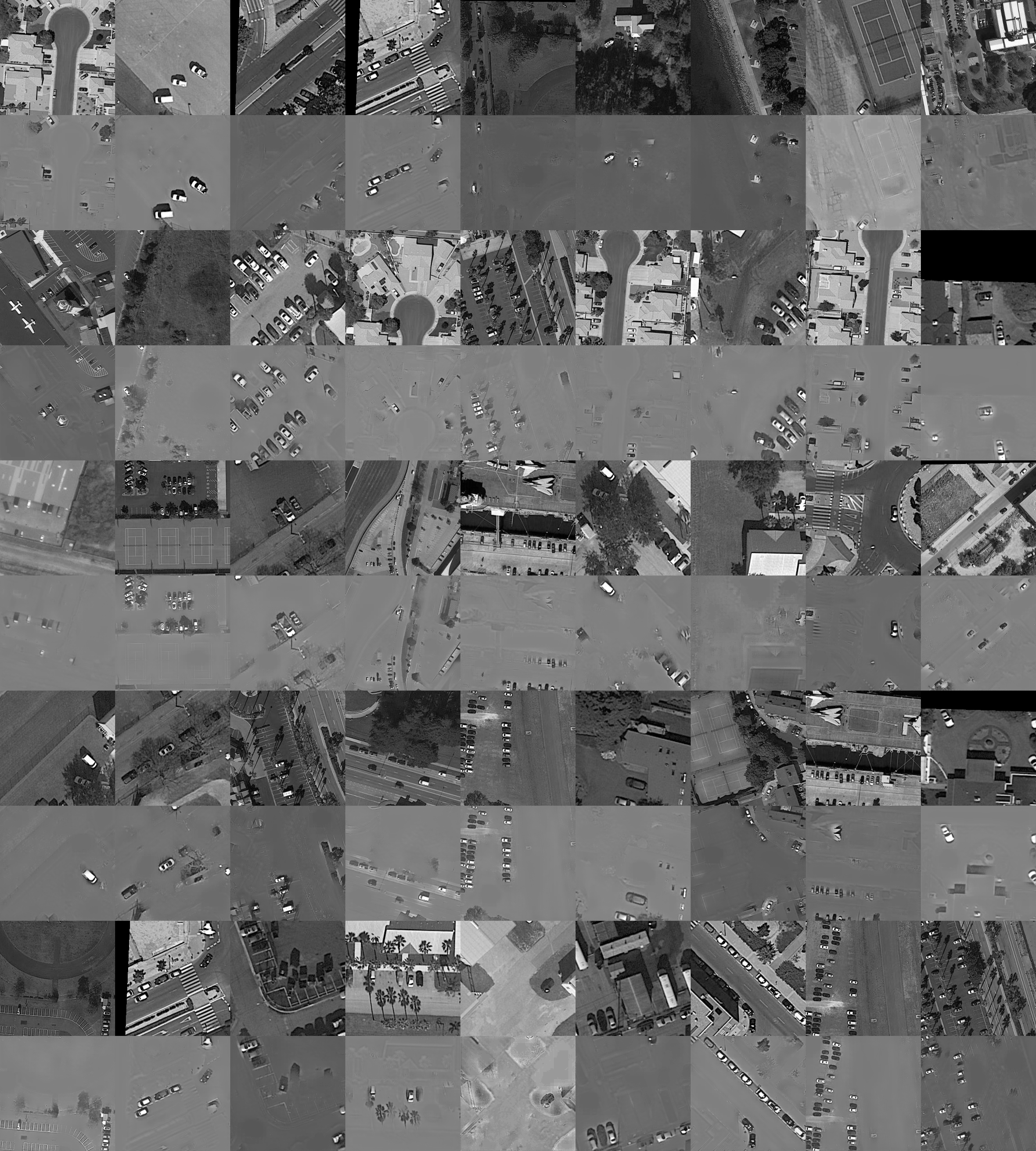}};
        \node(x_1)[left of=img1, rotate=0, xshift=-0.5\textwidth, yshift=0.5\textwidth] {\large $x_{orig}$};
        
        \node(psi1_1)[below of=x_1, rotate=0, yshift=-\yspace] {\large $\psi_1$};
        \node(x_2)[below of=psi1_1, rotate=0, yshift=-\yspace] {\large $x_{orig}$};
        \node(psi1_2)[below of=x_2, rotate=0, yshift=-\yspace] {\large $\psi_1$};
        \node(x_3)[below of=psi1_2, rotate=0, yshift=-\yspace] {\large $x_{orig}$};
        \node(psi1_3)[below of=x_3, rotate=0, yshift=-\yspace] {\large $\psi_1$};
        \node(x_4)[below of=psi1_3, rotate=0, yshift=-\yspace] {\large $x_{orig}$};
        \node(psi1_4)[below of=x_4, rotate=0, yshift=-\yspace] {\large $\psi_1$};
        \node(x_5)[below of=psi1_4, rotate=0, yshift=-\yspace] {\large $x_{orig}$};
        \node(psi1_5)[below of=x_5, rotate=0, yshift=-\yspace] {\large $\psi_1$};
        
        \draw[red, thick] (-0.5\textwidth,-0.332\textwidth) rectangle (0.5\textwidth,-0.222\textwidth);
        \draw[red, thick] (-0.5\textwidth,-0.11\textwidth) rectangle (0.5\textwidth,0\textwidth);
        \draw[red, thick] (-0.5\textwidth,0.11\textwidth) rectangle (0.5\textwidth,0.22\textwidth);
        \draw[red, thick] (-0.5\textwidth, 0.333\textwidth) rectangle (0.5\textwidth, 0.443\textwidth);
        \draw[red, thick] (-0.5\textwidth,-0.443\textwidth) rectangle (0.5\textwidth,-0.555\textwidth);
    \end{tikzpicture}
    }
    \caption{Additional results of cars  identification.}
    \label{fig:cars_table}
\end{figure}
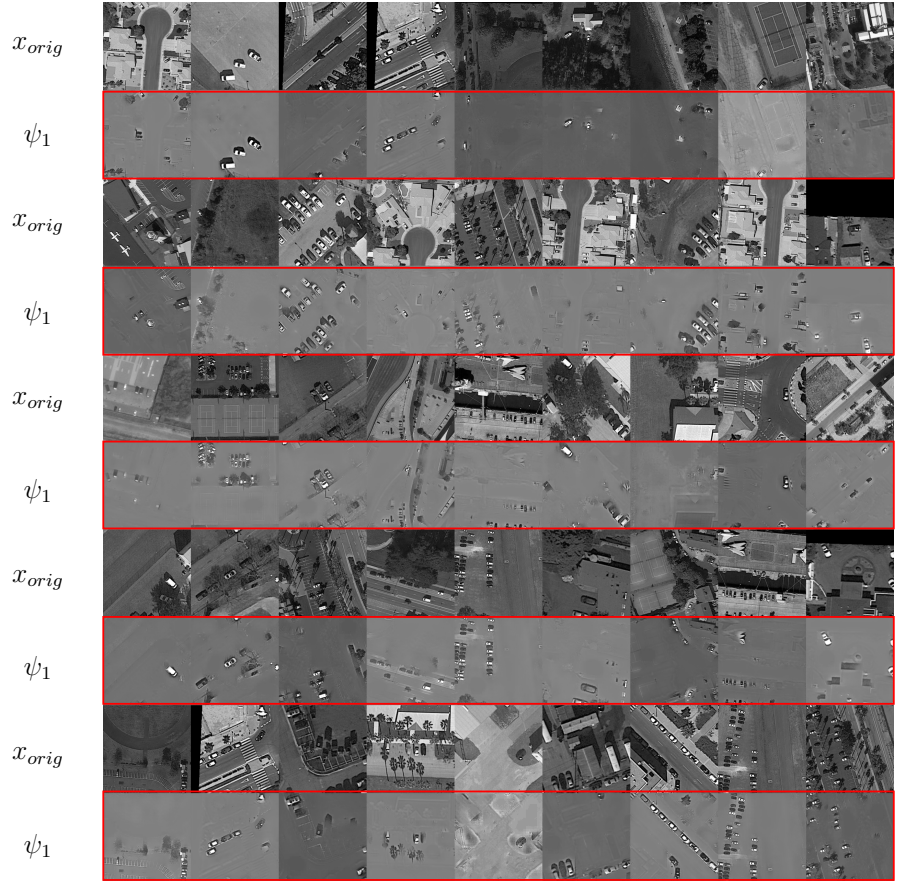


\bibliographystyle{plain}
\bibliography{main_branches.bib}
\end{document}